\documentclass{article}
\usepackage{arxiv}

\usepackage[utf8]{inputenc} 
\usepackage[T1]{fontenc}    
\usepackage{hyperref}       
\hypersetup{
    colorlinks,
    citecolor=blue,
    filecolor=black,
    linkcolor=blue,
    urlcolor=black
}
\usepackage{url}            
\usepackage{booktabs}       
\usepackage{amsmath,amssymb,amsfonts}       
\usepackage{nicefrac}       
\usepackage{microtype}      
\usepackage{cleveref}       
\usepackage{lipsum}         
\usepackage{graphicx}
\usepackage{doi}

\usepackage{IEEEtrantools}
\usepackage{soul}
\usepackage{cite}
\usepackage{float}
\usepackage{caption}
\usepackage{subcaption}
\usepackage{stfloats}
\usepackage{enumitem}
\usepackage{mathtools}
\usepackage{algorithm}
\usepackage{algpseudocode}
\usepackage{graphicx,color}
\usepackage{textcomp}
\usepackage{xcolor}
\usepackage{bm}
\usepackage{diagbox}
\usepackage{multirow}
\usepackage{makecell}
\usepackage{cases}
\usepackage{tikz}
\usepackage[british]{babel}
\usepackage[mathscr]{eucal}

\title{Incorporating Nonlocal Traffic Flow Model in Physics-informed Neural Networks}

\date{}

\author{
\href{https://orcid.org/0000-0001-6736-5627}{\includegraphics[scale=0.06]{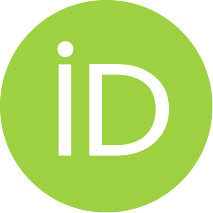}\hspace{1mm}Archie J.~Huang}\\
Department of Civil, Environmental \\
\& Construction Engineering\\
University of Central Florida\\
Orlando, FL, USA \\
\texttt{archie.huang@nyu.edu} \\
\And
\href{https://orcid.org/0000-0001-9544-3100}{\includegraphics[scale=0.06]{figures/orcid.pdf}\hspace{1mm}Animesh~Biswas}\\
Department of Mathematics\\
University of Nebraska-Lincoln\\
Lincoln, NE, USA\\
\texttt{abiswas2@unl.edu} \\
\And
\href{https://orcid.org/0000-0001-7754-6341}{\includegraphics[scale=0.06]{figures/orcid.pdf}\hspace{1mm}Shaurya~Agarwal}\\ 
Department of Civil, Environmental \\
\& Construction Engineering\\
University of Central Florida\\
Orlando, FL, USA \\
\texttt{shaurya.agarwal@ucf.edu} \\
}

\renewcommand{\shorttitle}{Incorporating Nonlocal Traffic Flow Model in Physics-informed Neural Networks}

\hypersetup{
pdftitle={Incorporating Nonlocal Traffic Flow Model in Physics-informed Neural Networks},
pdfsubject={cs.LG},
pdfauthor={Archie J.~Huang, Animesh~Biswas, Shaurya~Agarwal},
pdfkeywords={Physics Informed Machine Learning, Traffic State Estimation, Nonlocal Traffic Flow Model},
}

\begin{document}
\maketitle

\begin{abstract}
This research contributes to the advancement of traffic state estimation methods by leveraging the benefits of the nonlocal LWR model within a physics-informed deep learning framework. The classical LWR model, while useful, falls short of accurately representing real-world traffic flows. The nonlocal LWR model addresses this limitation by considering the speed as a weighted mean of the downstream traffic density. In this paper, we propose a novel PIDL framework that incorporates the nonlocal LWR model. We introduce both fixed-length and variable-length kernels and develop the required mathematics. The proposed PIDL framework undergoes a comprehensive evaluation, including various convolutional kernels and look-ahead windows, using data from the NGSIM and CitySim datasets. The results demonstrate improvements over the baseline PIDL approach using the local LWR model.  The findings highlight the potential of the proposed approach to enhance the accuracy and reliability of traffic state estimation, enabling more effective traffic management strategies.
\end{abstract}

\section{Introduction}
The paramount significance of traffic state estimation (TSE) for effective traffic management and control cannot be overstated  \cite{seo2017traffic, agarwal2019controllability}. Practitioners in transportation planning and engineering require a precise understanding of traffic flow conditions to implement various active traffic management and control (ATMC) measures such as varying-speed-limit \cite{yu2018varying} and freeway ramp-control \cite{agarwal2015feedback}. The advent and maturation of physics-informed deep learning (PIDL) \cite{raissi2019physics, cuomo2022scientific} and its proposed adoption in TSE \cite{huang2020physics, shi2022physics, huang2022physics} point to a promising direction towards achieving precise state estimation through limited/noisy observations of traffic. PIDL, in the context of TSE, ingrains the optimization process of a deep learning neural network with the governing equations in flow conservation laws, utilizing the regularization effect of the physics-cost term for improved traffic  estimation  \cite{di2023physics, huang2023limitations}. \cite{di2023physics} provides a comprehensive review of architectural designs of PIDL computational graphs and how these structures are customized to the TSE problem. Note that the current PIDL approaches have incorporated first-order classical Lighthill-Whitham-Richards (LWR) \cite{huang2020physics, huang2022physics, rempe2021estimating, shi2022physics}, second-order LWR models \cite{shi2022physics, zhao2022integrating, huang2023limitations}, and the discretized cell transmission model (CTM) \cite{huang2022physics}.

The classical LWR model, while pioneering, has revealed certain shortcomings; it predicates the average traffic speed as a function of traffic density, an assumption that falters under congested regimes \cite{chiarello2021overview}. Second-order models, comprising a mass conservation equation for density and an acceleration balance law for speed, overcome this limitation to some extent. Recent research has made progress in expanding the classical conservation-law-based traffic flow models to the \textbf{``nonlocal'' versions} \cite{abreu2022lagrangian, chiarello2021overview, bressan2020traffic}, wherein speed is conceptualized as a weighted mean of the downstream traffic density. This novel interpretation establishes speed as a Lipschitz function with respect to both space and time variables, ensuring bounded acceleration \cite{blandin2016well, sopasakis2006stochastic}. This effectively surmounts the constraints of classical macroscopic models that permitted speed discontinuities. In addition to transportation, nonlocal models have attracted sizable interest from experts in  mathematics and engineering due to their ability to capture multiple scales of interactions. These models find applications in different physical phenomena such as semi-permeable membranes problems in cellular biology \cite{biswas2021harnack}, nonlocal curvature and perimeters in image processing \cite{biswas2022nonlocal}, and dynamic fracture \cite{silling2000reformulation}.

\textbf{Contributions:} This paper advances the field of traffic state estimation (TSE), by incorporating the nonlocal LWR conservation law into a physics-informed deep learning (PIDL) paradigm. Specifically, the paper (a) demonstrates a novel methodology to incorporate nonlocal traffic flow into a PIDL neural network; (b) incorporates nonlocal density (as a new variable) using Greenshields fundamental diagram into the classical LWR model; (c) recasts the non-local LWR into an integro-differential equation - better suited for many PIDL architectures; (d) introduces two fixed length look ahead kernels and develop the mathematical framework for their application; (e) formulates a variable-length kernel overcoming the issue of ``thick'' upper boundary at the end of the road segment; (f) investigates empirically the effect of ``look-ahead'' window size. The validation with field data (NGSIM\cite{ngsim2021} and CitySim \cite{zheng2022citysim}) demonstrates the improvement in TSE results.

\textbf{Outline:} Section \ref{sec:classical} provides a brief background on the classical (local) LWR model and Greenshields fundamental diagram. Section \ref{sec:nonlocal} incorporates the nonlocal speed into the LWR model and recasts it as an integro-differential equation. Section \ref{sec:kernel} introduces fixed and variable length kernels. Section \ref{sec:archi} elaborates on the proposed PIDL architecture. Section \ref{sec:eval} evaluates the performance of the proposed PIDL and kernel functions using realistic data, and section \ref{sec:conc} provides concluding remarks.
%
%
\section{Classical Traffic Flow Model} \label{sec:classical}
%
This section provides a brief background on the classical (local) LWR model and Greenshields fundamental diagram. The Lighthill-Whitham-Richards (LWR) conservation law in the Eulerian coordinate system with the spatiotemporal location $\mathbf{X} = (x; \, t)$ as the independent variable is 
\begin{equation} \label{eqn:lwr_conservation_recall}
For \,\, (x, t) \, \in \, \mathbb{R} \, \times \, \mathbb{R}^{+} \, :   \,\,\,  \frac{\partial q(x, \, t)}{\partial x} + \frac{\partial \rho(x, \, t)}{\partial t} = 0
\end{equation}
Recall that fundamental diagrams establish the relationship between traffic state variables. The Greenshields' fundamental diagram (FD) which, among the traffic state variables, depicts the linear relationship between the velocity and density is given in \eqref{eqn:greenshields_fundamental_diagram_recall}. The jam-density is denoted as $\rho_m$, and the free-flow speed is $v_f$:
\begin{equation} \label{eqn:greenshields_fundamental_diagram_recall}
    \left\{ \,
        \begin{IEEEeqnarraybox}
            [\IEEEeqnarraystrutmode
            \IEEEeqnarraystrutsizeadd{7pt}{7pt}][c]{rCl}      
            q(x, \, t) & = & \rho(x, \, t) \, v_f \left(1 - \frac{\rho(x, \, t)}{\rho_m}\right)
            \\[2pt]
            v(x, \, t)  & = & v_f \left(1 - \frac{\rho(x, \, t)}{\rho_m} \right)
        \end{IEEEeqnarraybox}
    \right.
\end{equation}
Pairing the LWR partial differential equations (PDEs) and the Greenshields FD, we will have two ways to express the flow conservation law. Firstly, using the density variable $\rho(x, \, t)$ as the sole independent traffic state variable, we have  \eqref{eqn:lwr_conservation_density_recall}; and secondly, with velocity variable $v(x, \, t)$, we have \eqref{eqn:lwr_conservation_velocity_recall}.

\begin{equation} \label{eqn:lwr_conservation_density_recall}
    v_f \left(1 - \frac{2\rho(x, t)}{\rho_m}\right)\frac{\partial \rho(x, t)}{\partial x} + \frac{\partial \rho(x, t)}{\partial t} = 0
\end{equation}

\begin{equation} \label{eqn:lwr_conservation_velocity_recall}
    \rho_m \left(1 - \frac{2v(x, \; t)}{v_f}\right)  \frac{\partial v(x, \; t)}{\partial x}   - \frac{\rho_m}{v_f} \frac{\partial v(x, \; t)}{\partial t} = 0
\end{equation}

The incorporation of the local LWR model (Eqns \eqref{eqn:lwr_conservation_density_recall} and  \eqref{eqn:lwr_conservation_velocity_recall}) in PIDL and the resultant limitations were demonstrated in our previous works \cite{huang2020physics, huang2022physics} and \cite{huang2023limitations} respectively.
%
%
\section{Nonlocal Traffic Flow Model} \label{sec:nonlocal}
This section incorporates the nonlocal speed into the LWR model and recasts it as an integro-differential equation.

\begin{figure}[htbp]
\centerline{\includegraphics[width=0.45\textwidth]{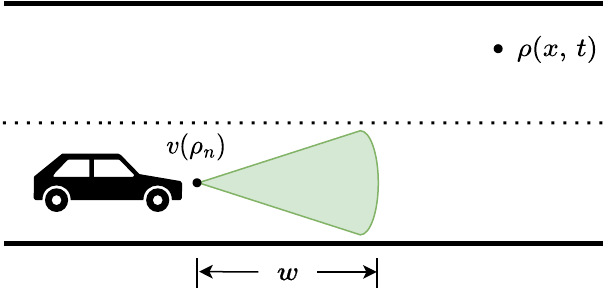}}
    \caption{Nonlocal Traffic States}
    \label{fig:nonlocal_traffic_states}
\end{figure}

Notice that through the fundamental diagram (e.g., Eqn \eqref{eqn:greenshields_fundamental_diagram_recall}), the traveling speed $v(x, \, t)$ at location $x$ and time $t$ can be determined by the local density $\rho(x, \, t)$. Therefore $v(x, \, t)$ can also be expressed as $v(\rho(x, \, t))$. The nonlocal version of traffic speed  $v(\rho_{n})$ at location $x$ and time $t$ is influenced by a representative nonlocal density $\rho_{n}$, which is obtained by a convolution of density values in the look-ahead window. The scenario is illustrated in Fig.~\ref{fig:nonlocal_traffic_states} where the green cone represents the look-ahead window, and the instantaneous speed of the vehicle is dependent upon the representative density in the look-ahead window. The length of the look-ahead window is denoted as $w$. The mathematical formulation is presented in \eqref{eqn:nonlocal_traffic_states}, where $\theta$ represents the kernel function.

\begin{equation} \label{eqn:nonlocal_traffic_states}
    \left\{ \,
        \begin{IEEEeqnarraybox}
            [\IEEEeqnarraystrutmode
            \IEEEeqnarraystrutsizeadd{7pt}{7pt}][c]{rCl}
            \rho_{n}(x, \, t) & = & \rho(x, \, t) \ast \theta (x) = 
            \int_{x}^{x + w} \rho(\tau, \, t) \theta (\tau - x) d\tau \\[2pt]
            v(x, \, t) & = & v(\rho_{n}(x, \, t))
        \end{IEEEeqnarraybox}
    \right.
\end{equation}

The Greenshields’ FD in \eqref{eqn:greenshields_fundamental_diagram_recall} is subsequently modified to reflect the relationship between the local velocity and the nonlocal density, as in \eqref{eqn:greenshields_fundamental_diagram_nonlocal}. Observe that in \eqref{eqn:greenshields_fundamental_diagram_nonlocal}, the local velocity $v(x, \, t)$ has the linear relationship with the convoluted nonlocal density $\rho_{n}(x, \, t)$, instead of the local density $\rho(x, \, t)$. 

\begin{equation} \label{eqn:greenshields_fundamental_diagram_nonlocal}
    \left\{ \,
        \begin{IEEEeqnarraybox}
            [\IEEEeqnarraystrutmode
            \IEEEeqnarraystrutsizeadd{7pt}{7pt}][c]{rCl}      
            q(x, \, t) & = & \rho(x, \, t) \, v_f \left(1 - \frac{\rho_{n}(x, \, t)}{\rho_m}\right)
            \\[2pt]
            v(x, \, t)  & = & v_f \left(1 - \frac{\rho_{n}(x, \, t)}{\rho_m} \right)
        \end{IEEEeqnarraybox}
    \right.
\end{equation}

The LWR conservation law, when articulated in relation to nonlocal traffic states, is delineated in \eqref{eqn:lwr_conservation_density_nonlocal}. Note that the nonlocal density $\rho_{n}(x, \, t)$ and the local density $\rho(x, \, t)$ are treated as \textbf{two separate variables} in \eqref{eqn:lwr_conservation_density_nonlocal}. This distinction serves as a cardinal principle in the subsequent development of the PIDL architecture.  In contrast to \eqref{eqn:lwr_conservation_density_recall}, here the partial derivative terms with respect to $x$ are calculated separately. 

\begin{equation} \label{eqn:lwr_conservation_density_nonlocal}
\begin{aligned}
    \frac{\partial \rho(x, \, t)}{\partial t} + v_f\left(1 - \frac{\rho_n(x, \, t)}{\rho_m}\right)  \frac{\partial \rho(x, \, t)}{\partial x} 
    - \rho(x, \, t) \frac{v_f}{\rho_m} \frac{\partial \rho_n(x, \, t)}{\partial x} = 0
\end{aligned}
\end{equation}

Next, we cast the non-local LWR model in terms of local density $\rho(x, \, t)$ and kernel function $\theta$. Assuming that $\theta$ is a smooth function of the distance variable and using Leibniz integration rule, we can write,
\begin{align} \label{eqn:lwr_rho_theta}
\begin{split}
    \frac{\partial \rho_n}{\partial x}(x,t) = -\int^{x+w}_{x} \rho(\tau,t) \frac{\partial \theta} 
     {\partial x}(\tau-x) \, d \tau 
     + \rho(x+w,t) \theta(w) - \rho(x,t) \theta(0)
    \end{split}
\end{align}

On the other hand, applying change of variable, $\tau = y+x$, we see that
\begin{align}
\rho_n(x,t) = \int^w_0 \rho(y+x,t) \theta(y) \, dy.  
\end{align}
In that case, 
\begin{align}\label{eqn:nl_rho_deriv}
   \frac{\partial \rho_n}{\partial x} (x,t) = \int^w_0 \frac{\partial \rho}{\partial x}(y+x,t) \theta(y) dy. 
\end{align}
Using \eqref{eqn:nl_rho_deriv}, we have, 
\begin{align}
     \frac{\partial}{\partial x}\bigg( \rho(x,t) \rho_n(x,t) \bigg) 
    = \int^w_0 \big(\rho(x,t) \rho_x(y+x,t) + \rho_x(x,t) \rho(x+y,t)\big) \theta(y) \, dy
\end{align}
Furthermore, using the fact $\int^w_0 \theta(y) dy = 1$, we have
\begin{align}
\rho(x,t) = \int^w_0 \rho(x,t) \theta(y) dy, \rho_x(x,t) = \int^w_0 \rho_x(x,t) \theta(y) dy. 
\end{align}
This transforms \eqref{eqn:lwr_conservation_density_nonlocal} into,
\begin{align}\label{eqn:nl_rho_deriv_conservation_density}
\begin{split}
     \frac{\partial \rho(x, \, t)}{\partial t} + v_f \int^w_0  \bigg( \rho_x(x,t) 
     - \frac{1}{\rho_m}\big( \rho(x,t) \rho_x(y+x,t) 
     + \rho_x(x,t) \rho(x+y,t) \big) \bigg)\theta(y) \, dy =0
     \end{split}
\end{align}
Notice that unlike \eqref{eqn:lwr_conservation_density_nonlocal}, equation \eqref{eqn:nl_rho_deriv_conservation_density} is only comprising of one variable $\rho$, which eliminates the need for computing $\rho_n$ separately, which further helps in the implementation of PIDL with the single variable. In the next section, we will show that even for the variable length kernel, we can write down the conservation equation only in terms of $\rho$.

\section{Kernel Functions} \label{sec:kernel}
This section introduces fixed-length and variable-length kernel functions to calculate the convoluted density in the look-ahead window. We also develop the mathematical framework for their application in the PIDL framework.

\subsection{Fixed-length Kernel Functions}

Recall that in \eqref{eqn:nonlocal_traffic_states}, $\theta (x)$ represents the kernel function in the look-ahead window $w$ and has the following property

\begin{equation} \label{eqn:kernel_property}
    \int_{x_{0}}^{x_{0} + w} \theta(x)dx = 1
\end{equation}

Next, we introduce two specific instances of fixed-length kernel functions, namely the \textit{constant kernel} and the \textit{linearly decreasing kernel}. 

\begin{figure}[htbp]
    \hspace{-5mm}\centerline{\includegraphics[width=0.35\textwidth]{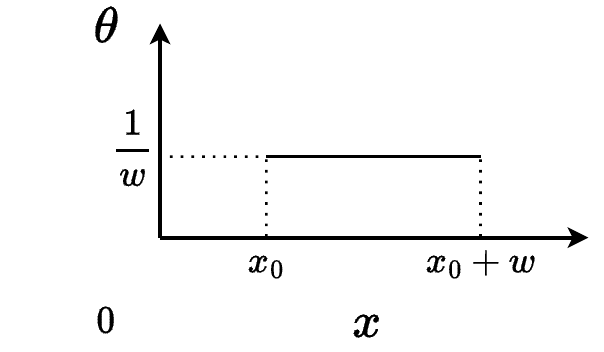}}
    \caption{The Constant Kernel}
    \label{fig:constant_kernel}
\end{figure}

The \textbf{{Constant Kernel}}   is designed to capture the mean downstream density within the driver's field of perception. Its mathematical formulation is given in \eqref{eqn:constant_kernel}, while its graphical representation is provided in Fig.~\ref{fig:constant_kernel}.

\begin{equation} \label{eqn:constant_kernel}
    \theta(x) = \frac{1}{w}, \;\;\; x \in [x_{0}, \; x_{0} + w]
\end{equation}

\begin{figure}[htbp]
    \hspace{-5mm}\centerline{\includegraphics[width=0.35\textwidth]{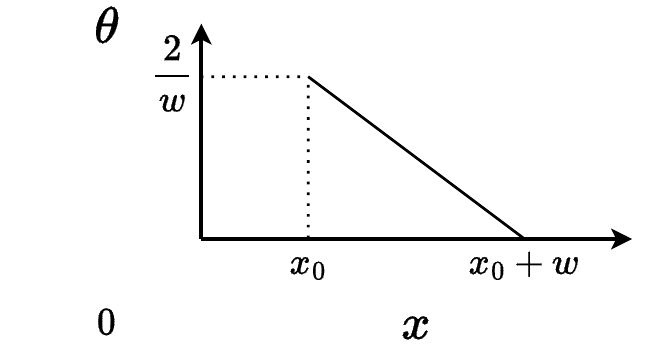}}
    \caption{The Linearly Decreasing Kernel}
    \label{fig:linearly_decreasing_kernel}
\end{figure}

The \textbf{Linearly Decreasing Kernel}, on the other hand, computes a weighted mean of density values within the driver's ``look-ahead'' zone. This kernel assigns a diminished weight to density values that are further from the driver's current position, thereby encapsulating the diminishing influence of distant traffic conditions. The formulation of this kernel is presented in \eqref{eqn:linearly_decreasing_kernel}, and Fig.~\ref{fig:linearly_decreasing_kernel} visually depicts the inverse relationship between the assigned weight and distance from the driver.

\begin{equation} \label{eqn:linearly_decreasing_kernel}
    \theta(x) = \frac{2}{w}(1 - \frac{x - x_{0}}{w}), \;\;\; x \in [x_{0}, \; x_{0} + w]
\end{equation}

\subsection{Variable-length Kernel}

\begin{figure}[htbp]
    \centerline{\includegraphics[width=0.7\textwidth]{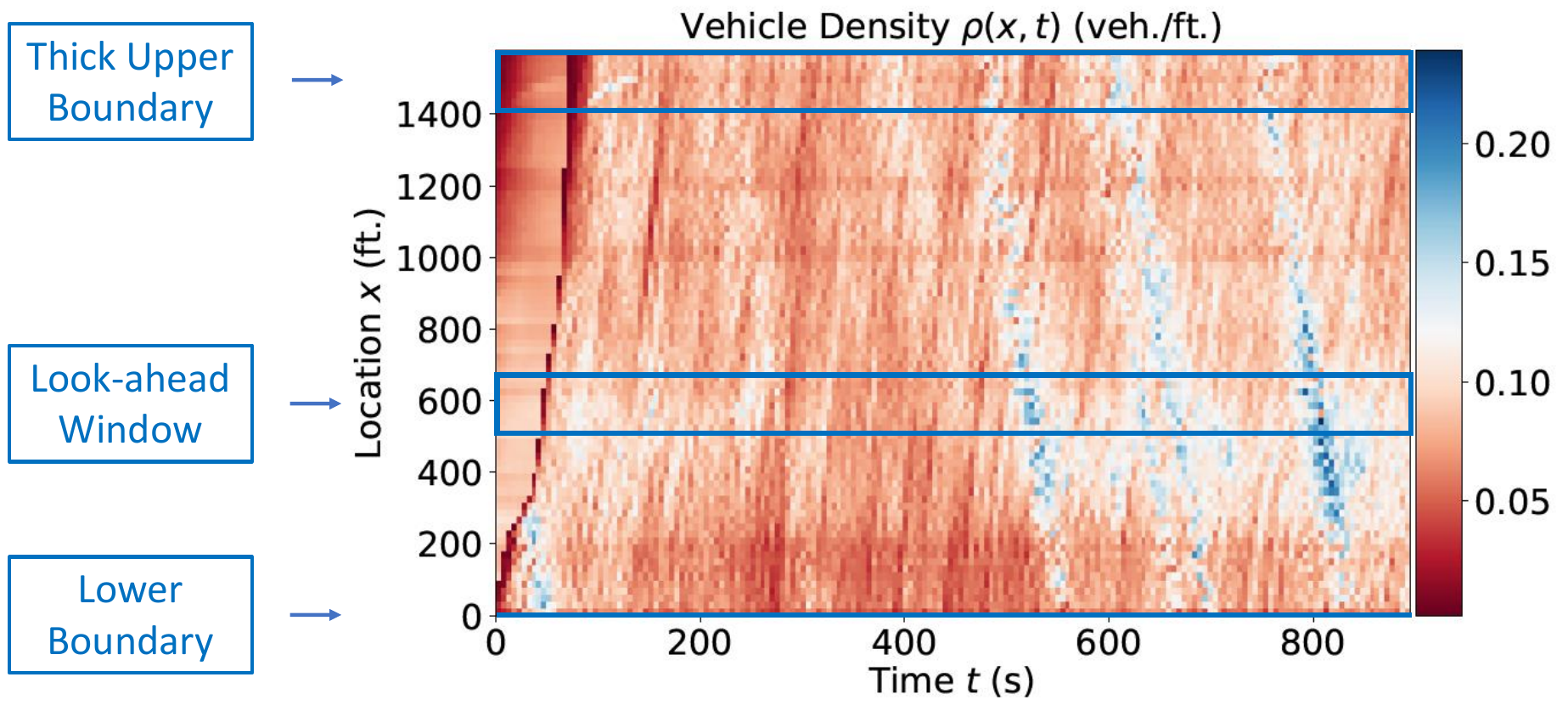}}
    \caption{``Thick'' Upper Boundary Condition Requirement}
    \label{fig:thick_boundary}
\end{figure}

The integration of nonlocal physics requires a ``thick'' upper boundary condition when the convolution window approaches the end of the road, see Fig.~\ref{fig:thick_boundary} for illustration.  With fixed-length kernels, the computation of nonlocal density values within this thick boundary is unattainable. 

In these circumstances, we can nonetheless employ the nonlocal model with a local boundary by utilizing a variable-length kernel \cite{tao2019nonlocal, silling2015variable}.  The premise behind the variable-length kernel is as follows: firstly we assume that the length of our fixed kernel is $w$ as in the previous case. Then consider a point $x$ inside our domain, we need to compute the distance of the point from the local boundary (thin boundary). If the distance is more than $w$, we can still use a fixed-length kernel of $w$-length to compute the nonlocal density at $x$. On the other hand, if the distance from the boundary is less than $w$, let us say $d_x$, then we use a $d_x$ length kernel (preserving the kernel's inherent nature at every point) to compute the nonlocal density at the point $x$. This method proves advantageous when data is exclusively available at the thin boundary. In numerous physical applications, collecting thick boundary data is impractical or unfeasible. For instance, consider density data at a traffic signal when the signal is red—it is logical to measure density at that specific location while gauging density data on the signal's right side is nonsensical.

The concept of variable-length kernel can be extrapolated to encompass every interior point within our domain, without strictly being restricted to the boundary vicinity. Here, we assume that the kernel length $w(x,t)$ at any given point $x$ is a function of local density, i.e., $w(x,t) = w(\rho(x,t))$. This model provides a closer representation of real-world scenarios. For instance, when the local density at a point $x$ is $\rho_m$, we would logically expect the velocity at that point to be $0$. However, if calculated using a fixed-length kernel, the nonlocal density at that point may not necessarily be $\rho_m$. This, in turn, implies that the velocity at that point isn't zero. To mitigate such discrepancies, we suggest employing a heterogeneous kernel, which we will introduce via a simple scheme below.

We define $w_0$ as the maximum length parameter, equivalent to the fixed length case ``$w$''. Next, we consider an interior point $x$, situated at least a distance $w_0$ from the upper or right boundary. Subsequently, we define: 

\begin{equation}
w(x,t) = w_0 \bigg(1 - \frac{\rho(x,t)}{\rho_m} \bigg)
\end{equation}

This equation indicates that the kernel length diminishes with an increase in local density. When the local density at any point is near its maximum, there is no requirement to consider local densities significantly distanced from that point. Conversely, when the local density at any point is near $0$, densities within a reasonable distance must be considered, as they may influence the velocity at that point. The nonlocal density at the point $(x_0,t)$ can therefore be computed as:
\begin{equation}
\rho_n(x_0,t) = \int^{x_0+w(x_0,t)}_{x_0} \rho(\tau,t) \theta_{(x_0,t)}(\tau-x_0) \, d\tau
\end{equation}

In the case of the linearly varying kernel, we have 
\begin{align}
    \theta_{(x_0,t)}(x) 
    = \frac{2}{w(x_0,t)} \big[1 - \frac{1}{w(x_0,t)} (x-x_0)\big] \chi_{(x_0, x_0 +w(x_0,t))}
\end{align}
where $\chi$ is the characteristics function.

For more general case, we still assume that $w(x,t) =0$ when $\rho(x,t) = \rho_m$ and $w(x,t)=w_0$ when $\rho(x,t) = 0$ and at any point $(x,t)$, the length of the kernel $w$ decreases if the density at that point increases. Let $\theta_0(y)$ be a smooth function in $[0, \infty)$ with the support set in $[0,w_0]$ such that $\int^\infty_0 \theta_0(y) dy = \int^{w_0}_0 \theta(y) dy = 1$. Notice that $\theta_0$ may not be necessarily smooth in $[0,\infty)$, for example, the constant and linearly varying kernel are not differentiable at $w_0$. For simplicity, we assume $\theta_0$ smoothly decreases to $0$ at $w_0$. We call $\theta_0$ to be the non-modulated kernel. Then we assume that $\theta_{(x,t)}$, the kernel to be used at any point $(x,t)$ is given by

\begin{equation}
\begin{cases}
    \theta_{(x,t)}(y) = \frac{w_0}{w(x,t)} \theta_0 \bigg(\frac{w_0}{w(x,t)} y \bigg), \quad \text{if}~w(x,t)>0 \\
    \theta_{(x,t)}(y) = \delta(y-x), \quad \text{if}~w(x,t)=0,
\end{cases}
\end{equation}
where $\delta(y-x)$ is the dirac delta function at $y=x$.
Then from the expression, it is evident that $\theta_{(x,t)}$ is supported in $[0,w(x,t)]$ and

\begin{align}
    \int^\infty_0 \theta_{(x,t)}(y) dy &= \int^{w(x,t)}_0 \theta_{(x,t)}(y) dy \nonumber \\
    &= \int^{w(x,t)}_0  \frac{w_0}{w(x,t)} \theta_0 \bigg(\frac{w_0}{w(x,t)} y \bigg) dy \nonumber  \\
    &= \int^{w_0}_0 \theta_0(y) dy =1
\end{align}

Then we have,
\begin{align}\label{eq:varying_kernel}
\rho_n(x,t) &= \int^{x+w(x,t)}_x \rho(\tau,t) \theta_{(x,t)}(\tau-x) \, d \tau \nonumber \\
&=\int^{w(x,t)}_0 \rho(x+\tau, t) \theta_{(x,t)}(\tau) \, d\tau \nonumber \\
&= \int^{w(x,t)}_0 \rho(x+\tau,t) \frac{w_0}{w(x,t)} \theta_0 \bigg(\frac{w_0}{w(x,t)} \tau \bigg) \, d \tau  \\
&= \int^{\infty}_0 \rho(x+\tau,t) \frac{w_0}{w(x,t)} \theta_0 \bigg(\frac{w_0}{w(x,t)} \tau \bigg) \, d \tau
\end{align}

The last equality is true since $\theta_0$ is supported in $[0,w_0]$.
Next we see that $w(x,t) = w(\rho(x,t))$ and we assume that $w$-varies smoothly with $\rho \in (0,\infty)$. Then taking derivative in the variable $x$, we have,

\begin{align}
    \frac{\partial \rho_n}{\partial x}(x,t) 
    = \int^\infty_0 \bigg[ \frac{\partial \rho}{\partial x}(x+\tau, t) \frac{w_0}{w(x,t)} \theta_0 \big(\frac{w_0}{w(x,t)} \tau \big) 
     + \rho(x+\tau,t) \frac{\partial}{\partial x} \bigg(\frac{w_0}{w(\rho(x,t))} \theta_0 \big(\frac{w_0}{w(\rho(x,t))} \tau \bigg)  \bigg] \, d\tau  
\end{align}

Now,
\begin{align}
    &\frac{\partial}{\partial x} \bigg(\frac{w_0}{w(\rho(x,t))} \theta_0 \big(\frac{w_0}{w(\rho(x,t))} \tau \big) \bigg)  \\
    &=\frac{\partial}{\partial \rho} \bigg(\frac{w_0}{w(\rho(x,t))} \theta_0 \big(\frac{w_0}{w(\rho(x,t))} \tau \big) \bigg) \frac{\partial \rho}{\partial x}  \nonumber \\
    &= - \frac{w_0 w'(\rho(x,t))}{w^2(\rho(x,t))}\bigg[\theta_0\big(\frac{w_0}{w(\rho(x,t))} \tau \big) \nonumber 
     + \frac{w_0 \tau}{w(\rho(x,t))} \theta'_0 \big(\frac{w_0}{w(\rho(x,t))} \tau \big) \bigg] \frac{\partial \rho}{\partial x} \nonumber
\end{align}
where $w', \theta'_0$ are the rate of change of $w$ and $\theta_0$ with respect to their arguments. Once we compute the derivatives we can write:

\begin{align} \label{eq:p_n deriv}
    &\frac{\partial \rho_n}{\partial x}(x,t) 
    = \int^{\infty}_0  \frac{\partial \rho}{\partial x}(x+\tau, t) \frac{w_0}{w(x,t)} \theta_0 \big(\frac{w_0}{w(x,t)} \tau \big) d\tau  \\
    &\quad \quad\quad\quad- \int^\infty_0 \rho(x+\tau,t) \frac{w_0 w'(\rho(x,t))}{w^2(\rho(x,t))}\bigg[\theta_0\big(\frac{w_0}{w(\rho(x,t))} \tau \big) \nonumber 
     + \frac{w_0 \tau}{w(\rho(x,t))} \theta'_0 \big(\frac{w_0}{w(\rho(x,t))} \tau \big) \bigg] \frac{\partial \rho}{\partial x} \, d\tau\nonumber
\end{align}

 Notice to prove the formulations mentioned above, we assume that $w$ varies with $\rho$ smoothly when $\rho \in (0,\infty)$ and $\theta_0$ varies with $y$ smoothly when $y \in (0,\infty)$. In fact using the smooth variation of $w$ and $\theta_0$ in their respective support sets, we can write \eqref{eq:p_n deriv} as
\begin{align}
    &\frac{\partial \rho_n}{\partial x}(x,t) 
    = \int^{w(x,t)}_0  \frac{\partial \rho}{\partial x}(x+\tau, t) \frac{w_0}{w(x,t)} \theta_0 \big(\frac{w_0}{w(x,t)} \tau \big) d\tau \\
    &\quad \quad\quad\quad- \int^{w(x,t)}_0 \rho(x+\tau,t) \frac{w_0 w'(\rho(x,t))}{w^2(\rho(x,t))}\bigg[\theta_0\big(\frac{w_0}{w(\rho(x,t))} \tau \big) \nonumber 
    + \frac{w_0 \tau}{w(\rho(x,t))} \theta'_0 \big(\frac{w_0}{w(\rho(x,t))} \tau \big) \bigg] \frac{\partial \rho}{\partial x}\, d\tau \nonumber
\end{align}
On the other hand, if these functions are not smooth in $[0,\infty)$, for example $\theta_0$ is constant or linear and $w(\rho)$ is linear, we can use \eqref{eq:varying_kernel} and do the change of variable to get,
$$\rho_n(x,t) = \int^{w_0}_0 \rho \big(x+ \frac{w(\rho(x,t))}{w_0} \tau, t \big) \theta_0 (\tau) \, d\tau.$$
Taking derivative in the variable $x$, we have
\begin{align}
    \frac{\partial \rho_n}{\partial x}(x,t)
    = \int^{w_0}_0 \frac{\partial \rho}{\partial x} \big(x+ \frac{w(\rho(x,t))}{w_0} \tau, t \big) \frac{\tau}{w_0} w'(\rho(x,t)) \frac{\partial \rho}{\partial x}(x,t) \, d\tau 
\end{align}
For the expression above, we assume that $w$ does not have any jump discontinuity at $\rho_m$ and $w$ is smooth in the region $(0, \rho_m)$. It may not be differentiable at $\rho_m$. On the other hand, $\theta_0$ does not need any regularity. 
Since we write $\rho_n(x,t)$ and $\frac{\partial \rho_n}{\partial x}(x,t)$ only in terms of $\rho(x,t), \rho_x(x,t)$ we can have the conservation equation only in terms of $\rho(x,t)$.
The above-discussed alternative mathematical representation of the kernel functions aids in incorporating them into various PIDL architectures.

\section{PIDL Architecture with Nonlocal Physics} \label{sec:archi}

\begin{figure}[h]
    \centerline{\includegraphics[width=0.7\textwidth]
    {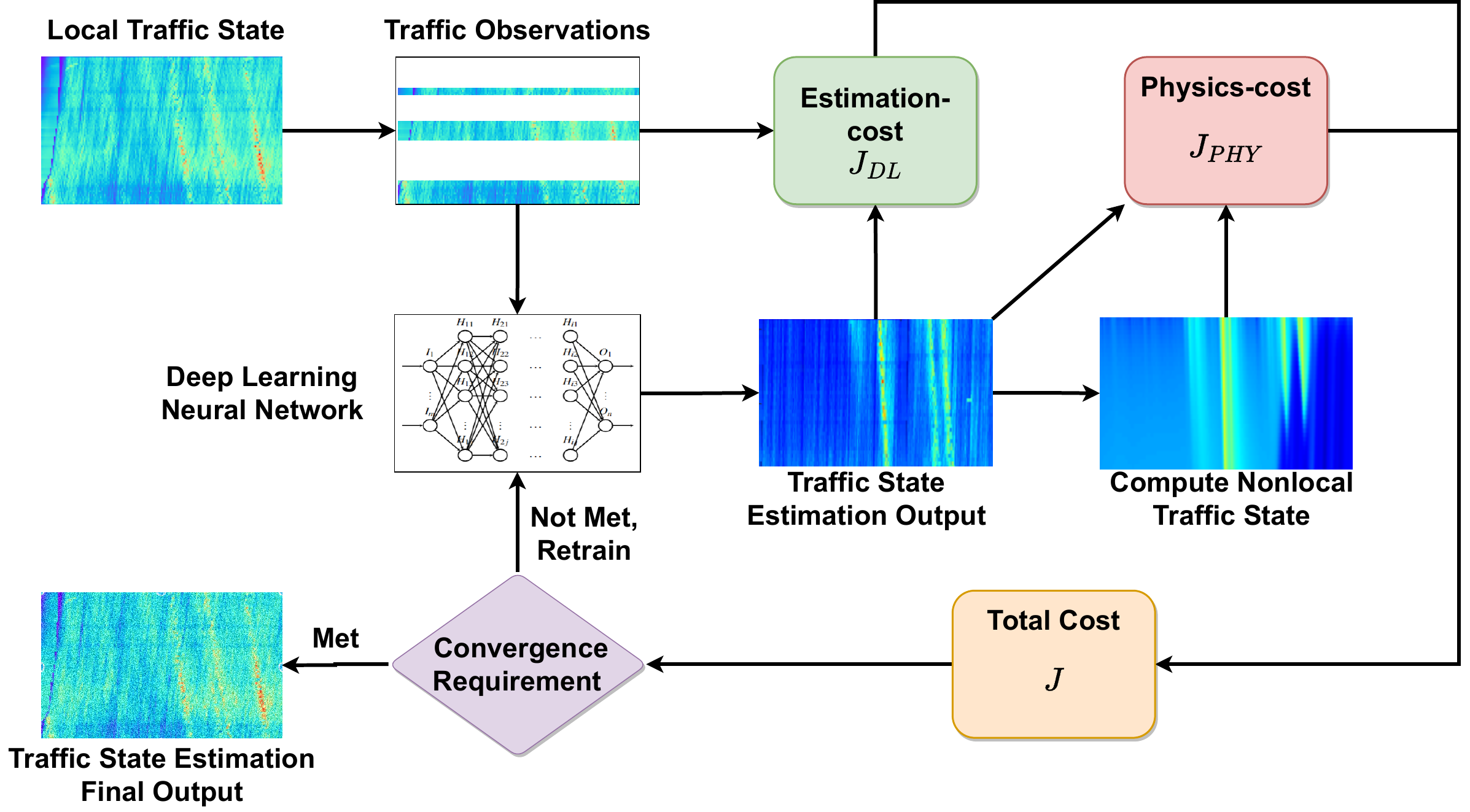}}
    \caption{Approach Overview of PIDL with Nonlocal Physics for TSE}
    \label{fig:approach_overview_pidl_nonlocal}
\end{figure}

Expanding upon our prior endeavors in physics-informed deep learning for traffic state estimation \cite{huang2022physics, huang2023limitations}, we present the development of a novel paradigm of PIDL incorporating nonlocal physics for TSE. The schematic of the overall methodology is illustrated in Fig.~\ref{fig:approach_overview_pidl_nonlocal}. In the course of the neural network optimization, a convolution kernel is applied to the provisional state estimation output to construct the field of nonlocal traffic states. This construction then assists in the computation of the physics-cost term $J_{PHY}$, subsequently leading to the derivation of the total cost $J$ of the provisional state estimation outcome. The PIDL process iteratively refines this output until the cost of the optimized output fulfills the specified convergence criterion. Upon satisfaction of this condition, the PIDL returns the refined state estimation as the final output. This iterative, physics-informed optimization strategy ensures a rigorous and nuanced traffic state estimation that encapsulates the nonlocal dynamics inherent in real-world traffic scenarios.

The algorithmic implementation of the PIDL model that integrates nonlocal physics principles is delineated in Algorithm \ref{alg:pidl_nonlocal}. The training inputs encompass density observations located at $\mathrm{P_{O}} = \{\rho(x_{o}^{j}, \, t_{o}^{j}) | j = 1, 2, \cdots, N_{o}\}$, and the set of collocation points is denoted as $\mathrm{C} = \{(x_{c}^{k}, t_{c}^{k}) | k = 1, 2, \cdots, N_{c}\}$, where the computation of the physics-cost term takes place. Employing the Mean Squared Error (MSE) as the measure of error, the deep learning cost $J_{DL}$ and the physics-cost $J_{PHY}$ can be expressed as in \eqref{eqn:error_metric_l2}. The total cost function is formulated as a weighted summation of these two cost terms, $J_{DL}$ and $J_{PHY}$, mediated by a parameter $\mu$: $J = \mu * J_{DL} + (1 - \mu) * J_{PHY}$. This balanced approach enables the algorithm to maintain a judicious compromise between the data fidelity and the physical fidelity, ensuring the efficacy of the proposed PIDL model. For a more comprehensive understanding of the nuances involved in the implementation of PIDL for TSE, we encourage readers to refer to  \cite{huang2020physics, shi2022physics, huang2022physics, di2023physics, huang2023limitations}. 

\begin{equation} \label{eqn:error_metric_l2}
    \left\{ \,
        \begin{IEEEeqnarraybox}
            [\IEEEeqnarraystrutmode
            \IEEEeqnarraystrutsizeadd{7pt}{7pt}][c]{rCl}      
            J_{DL} & = & \frac{1}{N_{o}}\sum_{j=1}^{N_{o}}\left|\rho(x_{o}^{j}, \, t_{o}^{j}) - \hat{\rho}(x_{o}^{j}, \, t_{o}^{j})\right|^2  
            \\[2pt]
            J_{PHY} & = & \frac{1}{N_{c}}\sum_{k=1}^{N_{c}} \bigg| v_f\left(1 - \frac{\hat{\rho}_n(x_{c}^{k}, \, t_{c}^{k})}{\rho_m}\right)  \frac{\partial \hat{\rho}(x_{c}^{k}, \, t_{c}^{k})}{\partial x}
             +  \frac{\partial \hat{\rho}(x_{c}^{k}, \, t_{c}^{k})}{\partial t} - \hat{\rho}(x_{c}^{k}, \, t_{c}^{k}) \frac{v_f}{\rho_m} \frac{\partial \hat{\rho}_n(x_{c}^{k}, \, t_{c}^{k})}{\partial x}\bigg|^2
        \end{IEEEeqnarraybox}
    \right.
\end{equation}

\begin{algorithm}
\caption{PIDL Optimization with Nonlocal Physics}\label{alg:pidl_nonlocal}
\begin{flushleft}
\begin{algorithmic}
    \Require Density observations $\mathrm{P_{O}} = \{\rho(x_{o}^{j}, \, t_{o}^{j}) | j = 1, 2, \cdots, N_{o}\}$
    \Require Collocation points $\mathrm{C} = \{(x_{c}^{k}, t_{c}^{k}) | k = 1, 2, \cdots, N_{c}\}$
    \State Initialize neural network $\Psi(\theta)$ 
    \While{Convergence not reached}
    \State Generate density estimation $\mathrm{\hat{P}}$
    \State Compute $J_{DL} = MSE_{(\rho(x, t), \, \hat{\rho}(x, t))}$   
    \State Compute $J_{PHY} = MSE_{(0, \, \frac{\partial \rho(x, \, t)}{\partial t} + v_f\left(1 - \frac{\rho_n(x, \, t)}{\rho_m}\right)  \frac{\partial \rho(x, \, t)}{\partial x} - \rho(x, \, t) \frac{v_f}{\rho_m} \frac{\partial \rho_n(x, \, t)}{\partial x})}$ 
    \State Compute $J = \mu * J_{DL} + (1 - \mu) * J_{PHY}$ 
    \If{Convergence reached}
        \State Output density estimation $\mathrm{\hat{P}}$
    \Else
        \State Retrain  $\Psi(\theta)$
    \EndIf
    \EndWhile
\end{algorithmic}
\end{flushleft}
\end{algorithm}

\section{Evaluation and Validation} \label{sec:eval}

This section utilizes traffic observations from the next generation simulation (NGSIM) \cite{ngsim2021} and CitySim \cite{zheng2022citysim} datasets to train the proposed nonlocal PIDL paradigm. Subsequently, we contrast the estimation outcomes of traffic density, produced by this paradigm, with those generated by a benchmark PIDL model that employs the local LWR conservation law. This comparative analysis allows us to evaluate the efficacy of the nonlocal PIDL approach in providing more accurate traffic density estimates.

\subsection{Validation with NGSIM Data}

\begin{figure}[!htbp]
    \centerline{\includegraphics[width=0.57\textwidth]{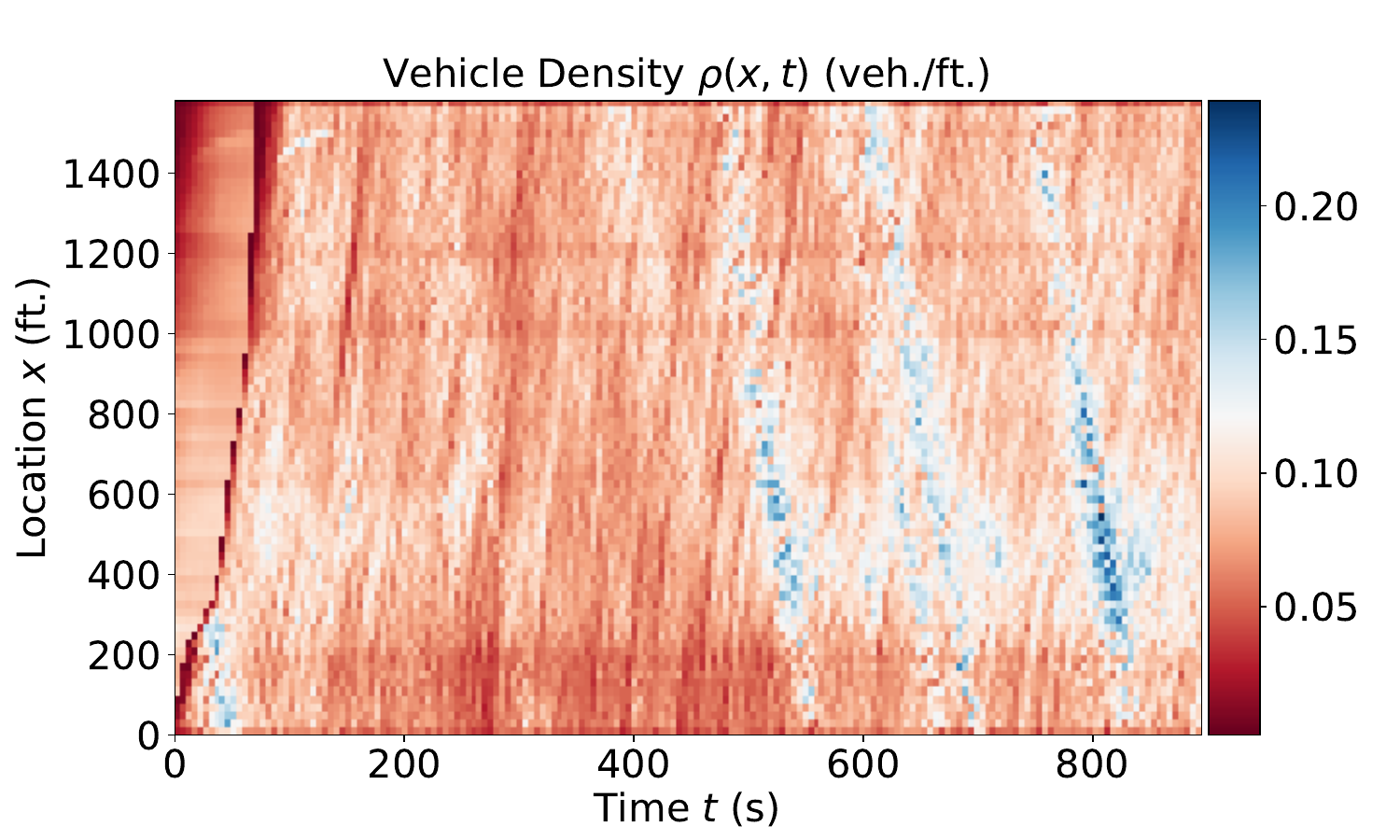}}
    \caption{NGSIM Vehicle Density-Field}
    \label{fig:ngsim_data_I80_4pm_density}
\end{figure}

We harness vehicle density data from the $I-80$ freeway in Emeryville, California, as procured from the NGSIM dataset \cite{ngsim2021}, to evaluate the effectiveness of PIDL with nonlocal physics. The chosen 15-minute snapshot of vehicular flow data, as represented in Fig.~\ref{fig:ngsim_data_I80_4pm_density}, was cataloged on April 13, 2005, between 4:00 pm and 4:15 pm.

\begin{figure*}[!bp]
     \centering
     \begin{subfigure}[b]{0.33\textwidth}
         \centering
         \includegraphics[width=\textwidth]{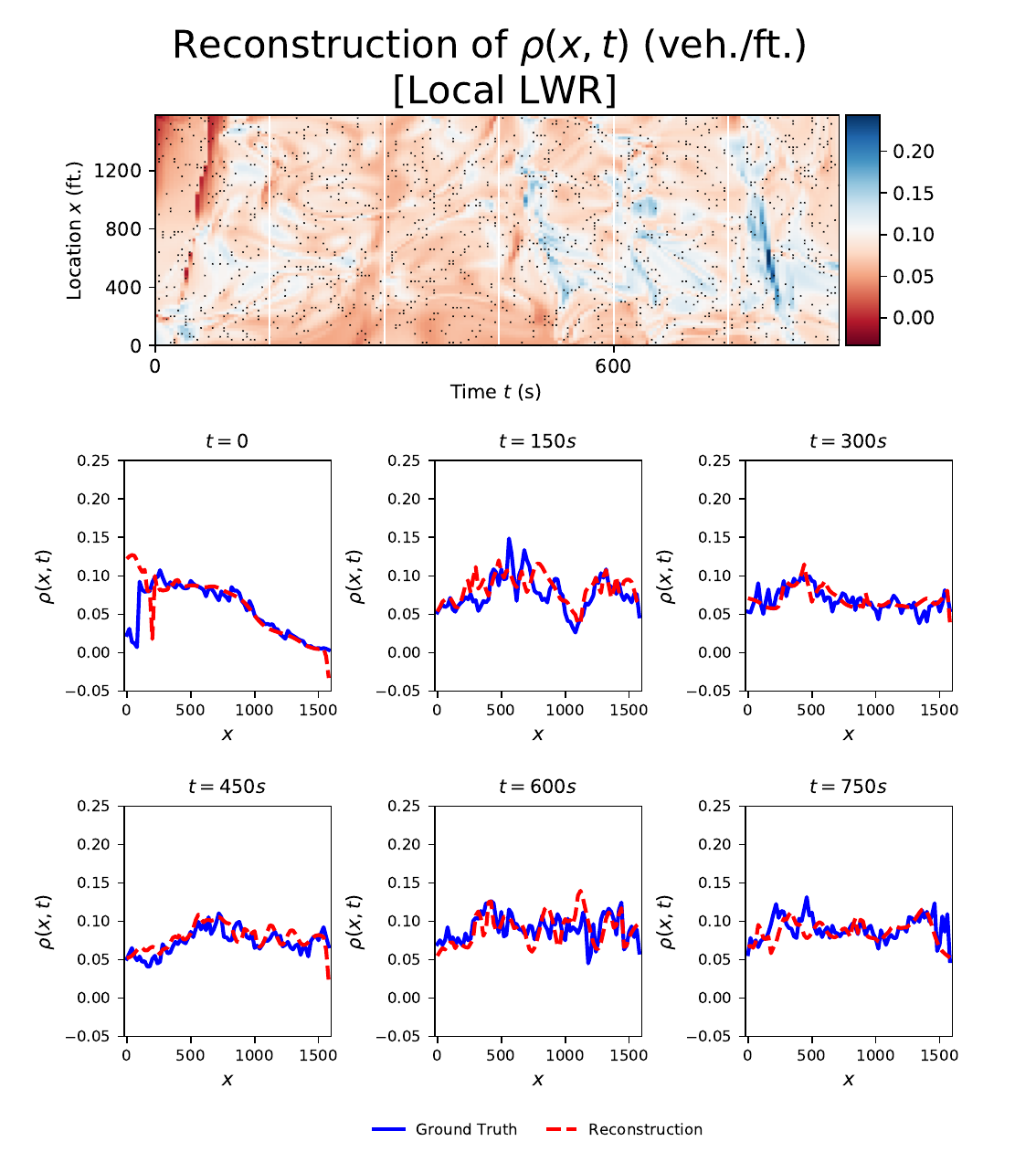}
    \caption{Local LWR}
    \label{fig:ngsim_local_kernel}
     \end{subfigure}
     \hfill
     \begin{subfigure}[b]{0.33\textwidth}
         \centering
         \includegraphics[width=\textwidth]{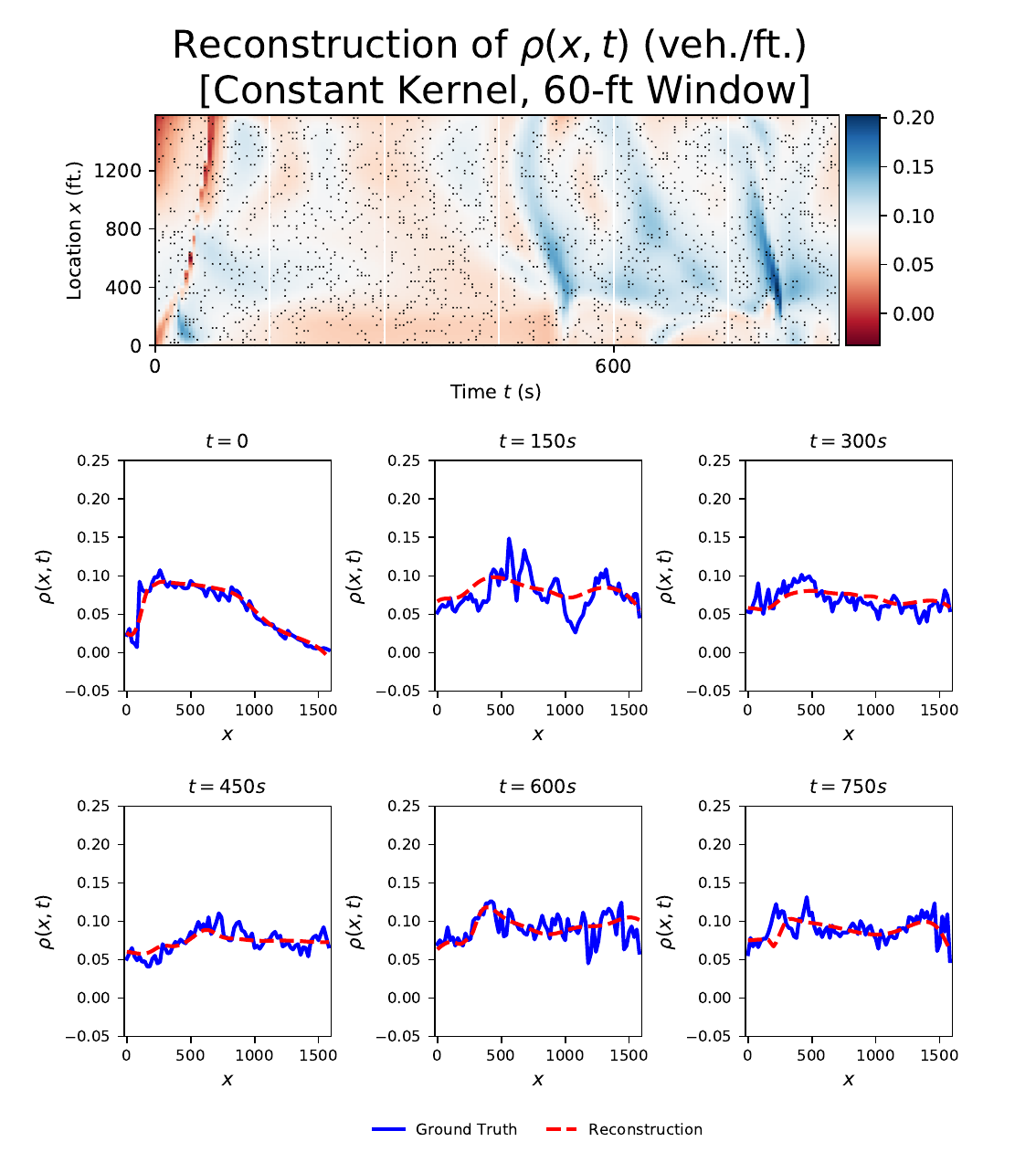}
    \caption{Constant Kernel, 60-ft}
    \label{fig:ngsim_constant_kernel_60}
     \end{subfigure}
    \hfill
     \begin{subfigure}[b]{0.33\textwidth}
         \centering
         \includegraphics[width=\textwidth]{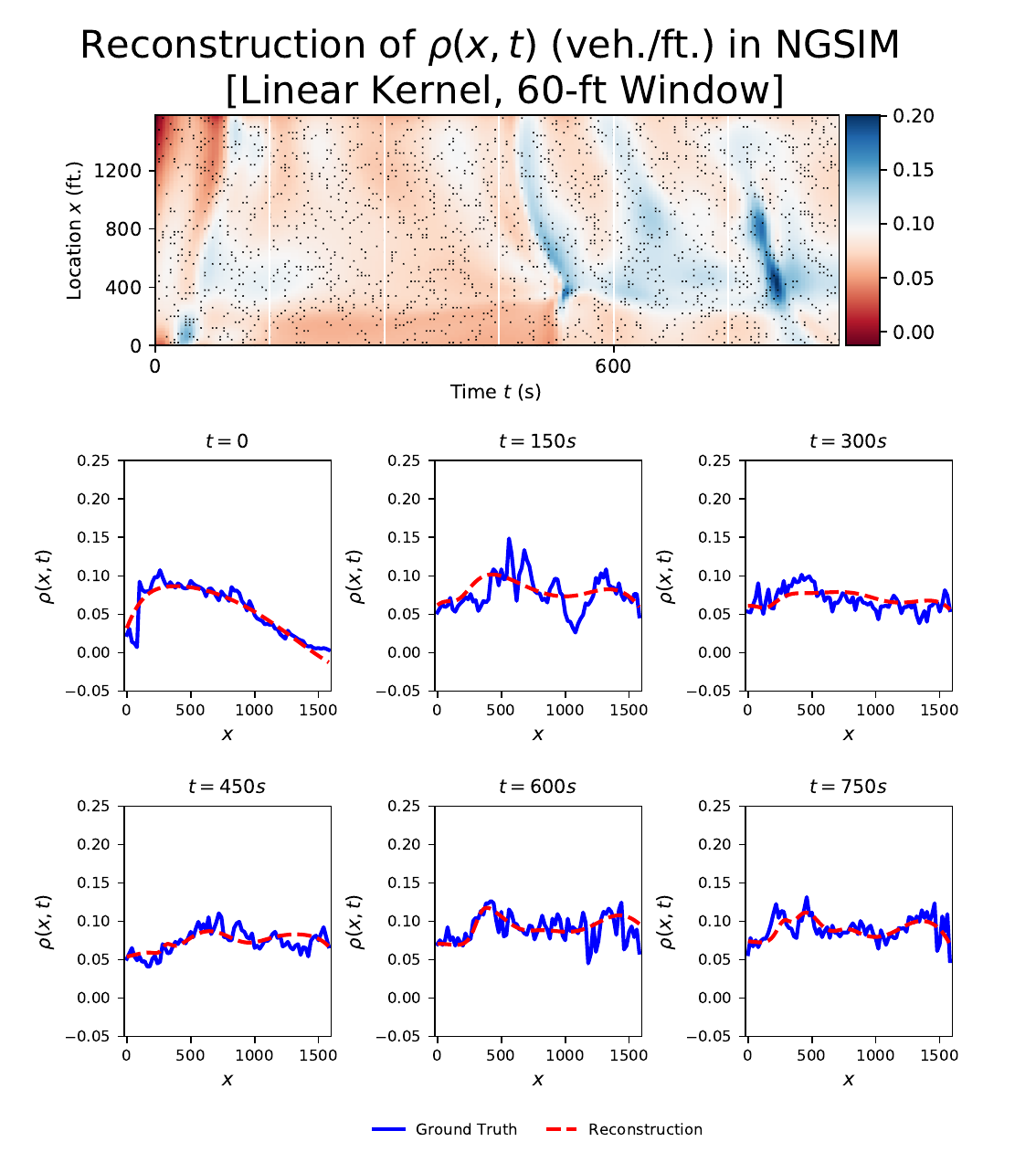}
    \caption{Linearly Decreasing Kernel, 60-ft}
    \label{fig:ngsim_linear_kernel_60}
     \end{subfigure}
    \caption{NGSIM Reconstruction with Local and Nonlocal LWR incorporated PIDL}
    \label{fig:ngsim_reconstruction}
\end{figure*}

We used two distinct kernel functions - the constant kernel and the linearly decreasing kernel, both  paired with a 60-foot lookahead window. The reconstruction outcomes are visually presented in Fig.~\ref{fig:ngsim_reconstruction}, while a tabulated summary of the relative $\mathcal{L}_2$ errors of the reconstructions can be found in Table \ref{tab:experiment_ngsim_I80_density_nonlocal}.

\begingroup
\setlength{\tabcolsep}{6pt} 
\renewcommand{\arraystretch}{1.4} 
    \begin{table*}[!htbp]
        \caption{Performance Analysis of Nonlocal PIDL using NGSIM Data}
        \begin{center} 
            \begin{tabular}{|p{0.5cm}|p{2.7cm}|p{3cm}|p{3.5cm}|p{3.5cm}|}\hline
                & \textbf{Model Type} & \textbf{Look-ahead Kernel} & \textbf{Convolution Window} & \textbf{Relative $\mathcal{L}_2$ Error (\%)}\\\hline
                \textbf{1}  &  Local LWR   & $N/A$ & $N/A$ & 20.70\\\hline
                \textbf{2}  &  Nonlocal LWR   & Constant & 60-ft & 17.65\\\hline
                \textbf{3}  &  Nonlocal LWR   & Linearly decreasing & 60-ft & \textbf{17.40} \\\hline
            \end{tabular}
            \label{tab:experiment_ngsim_I80_density_nonlocal}
        \end{center}
    \end{table*}
\endgroup

Compared to the reconstruction result of using local LWR (Fig.~\ref{fig:ngsim_local_kernel}), we observe the additional regularization effect with the use of nonlocal physics and incremental improvement in terms of reconstruction accuracy using the 60-ft look-ahead window constant kernel (Fig.~\ref{fig:ngsim_constant_kernel_60}) and the linearly decreasing kernel (Fig.~\ref{fig:ngsim_linear_kernel_60}). Notice that the incorporation of nonlocal physics into PIDL yields dual benefits. Firstly, it expedites the training process of the neural network, and  secondly, it ensures the preservation of shockwaves in the reconstruction.   

\subsection{Validation with CitySim Data}

We deploy the PIDL with nonlocal physics on the CitySim dataset \cite{zheng2022citysim}; it is a vehicle trajectory dataset extracted from drone videos on urban highways. Our chosen subset of the CitySim dataset encompasses a 300-second drone recording capturing the vehicular trajectories along a 2400-ft freeway segment. Importantly, this specific freeway segment was selected due to the absence of any on-ramps or off-ramps, thereby guaranteeing the conservation of traffic flow.

\begin{figure*}[htbp]
     \centering
     \begin{subfigure}[b]{0.33\textwidth}
         \centering
         \includegraphics[width=\textwidth]{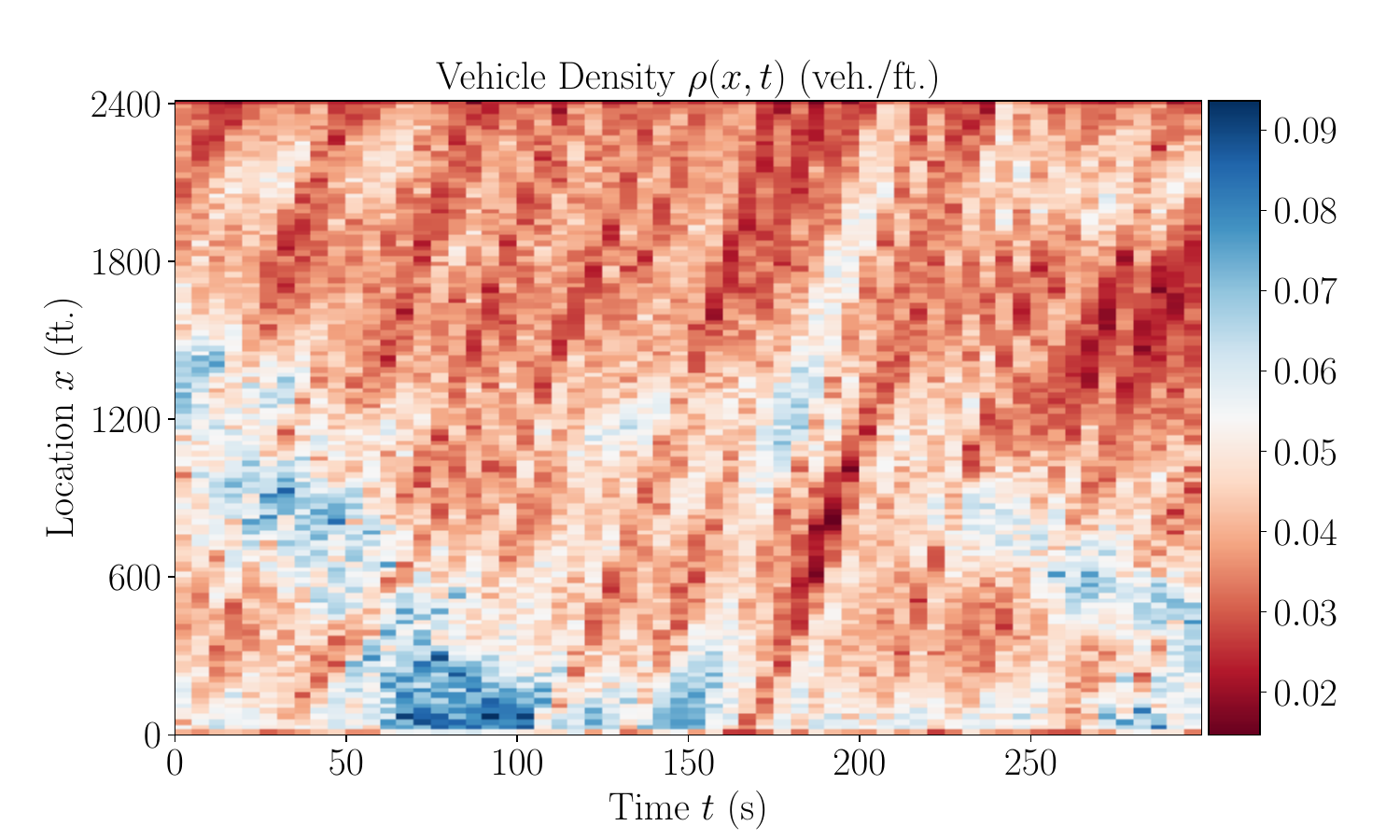}
        \caption{Density Field}
        \label{fig:citysim_density_data}
     \end{subfigure}
     \hfill
     \begin{subfigure}[b]{0.33\textwidth}
         \centering
         \includegraphics[width=\textwidth]{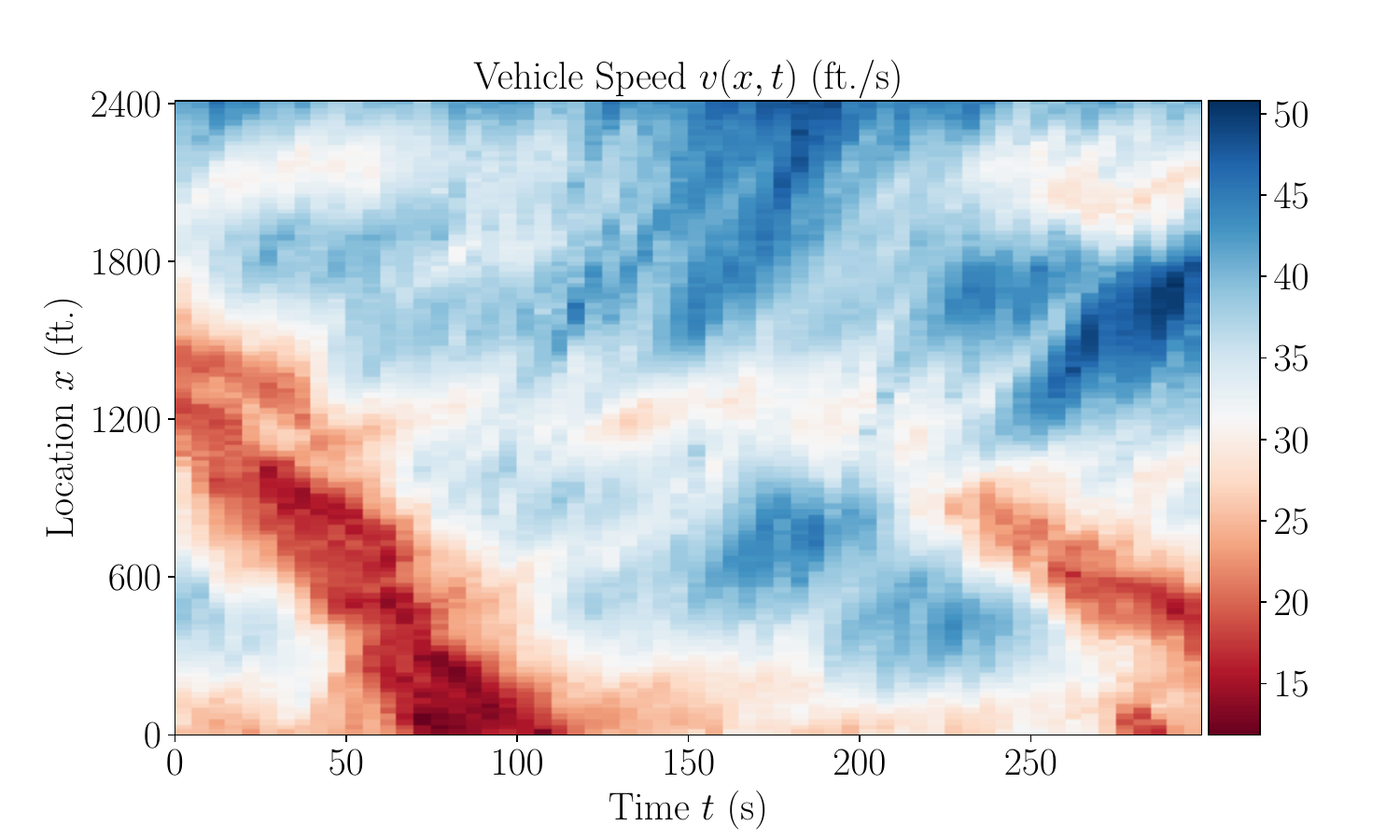}
    \caption{Velocity Field}
    \label{fig:citysim_velocity_data}
     \end{subfigure}
    \hfill
     \begin{subfigure}[b]{0.33\textwidth}
         \centering
             \includegraphics[width=\textwidth]{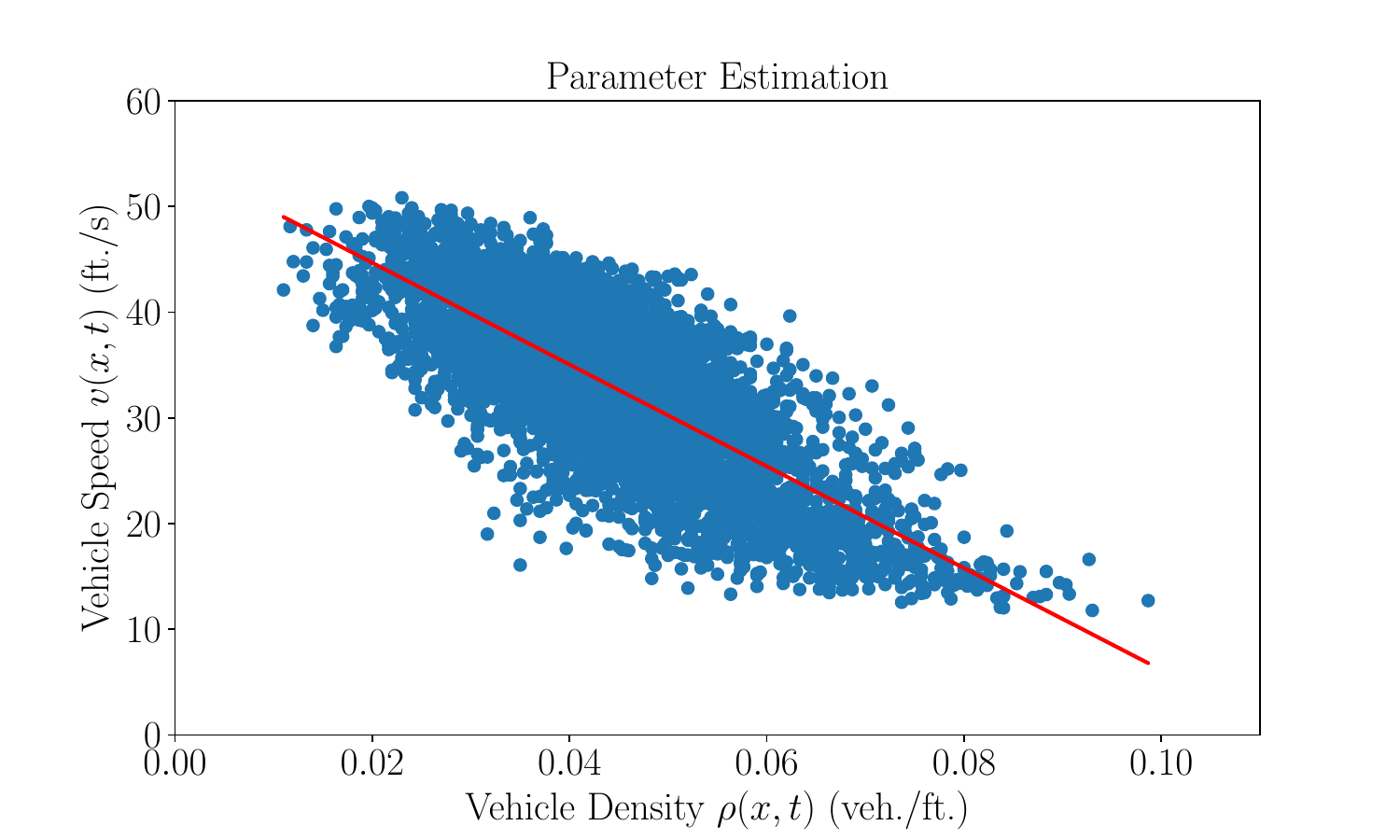}
    \caption{Parameter Estimation}
    \label{fig:citysim_parameter_estimation}
     \end{subfigure}
    \caption{CitySim: Traffic Fields and Parameter Estimation}
    \label{fig:citysim_data}
\end{figure*}

The recorded vehicle trajectories are aggregated in both spatial and temporal domains, using bins characterized by a length of 20 feet and a time interval of 5 seconds ($\Delta x = 20 ft., \Delta t = 5s$). The vehicle density, visualized in Fig.~\ref{fig:citysim_density_data}, represents an accumulation of vehicle densities across the three traffic lanes. Meanwhile, the vehicle speed averaged over the aggregated bins, is depicted in Fig.~\ref{fig:citysim_velocity_data}. Parameters for the dataset are estimated and presented in Fig.~\ref{fig:citysim_parameter_estimation}. The free-flow speed, $v_f$, is assessed at approximately 54.30 feet per second, whereas the jam density, $\rho_m$, is calculated to be around 0.11 vehicles per foot (a total across the three lanes).

We utilize 1000 traffic density points ($13.77\%$ of the entire dataset) as training observations and 5000 collocation points to compute physics cost. A comparative analysis of the relative $\mathcal{L}_2$ error, obtained through reconstruction using the local and nonlocal LWR (with constant kernels and linearly decreasing kernels of 60-ft and 100-ft look-ahead windows) is summarized in Table \ref{tab:citysim_I80_density_nonlocal} (Models 1 to 5). Aside from implementing the fixed-length kernels, we also incorporate the variable-length kernels with a maximum length of $100$ ft and tabulate the results in Table~\ref{tab:citysim_I80_density_nonlocal} (Models 6 and 7). 

\begin{figure*}[htbp]
     \centering
     \begin{subfigure}[b]{0.33\textwidth}
         \centering
         \includegraphics[width=\textwidth]{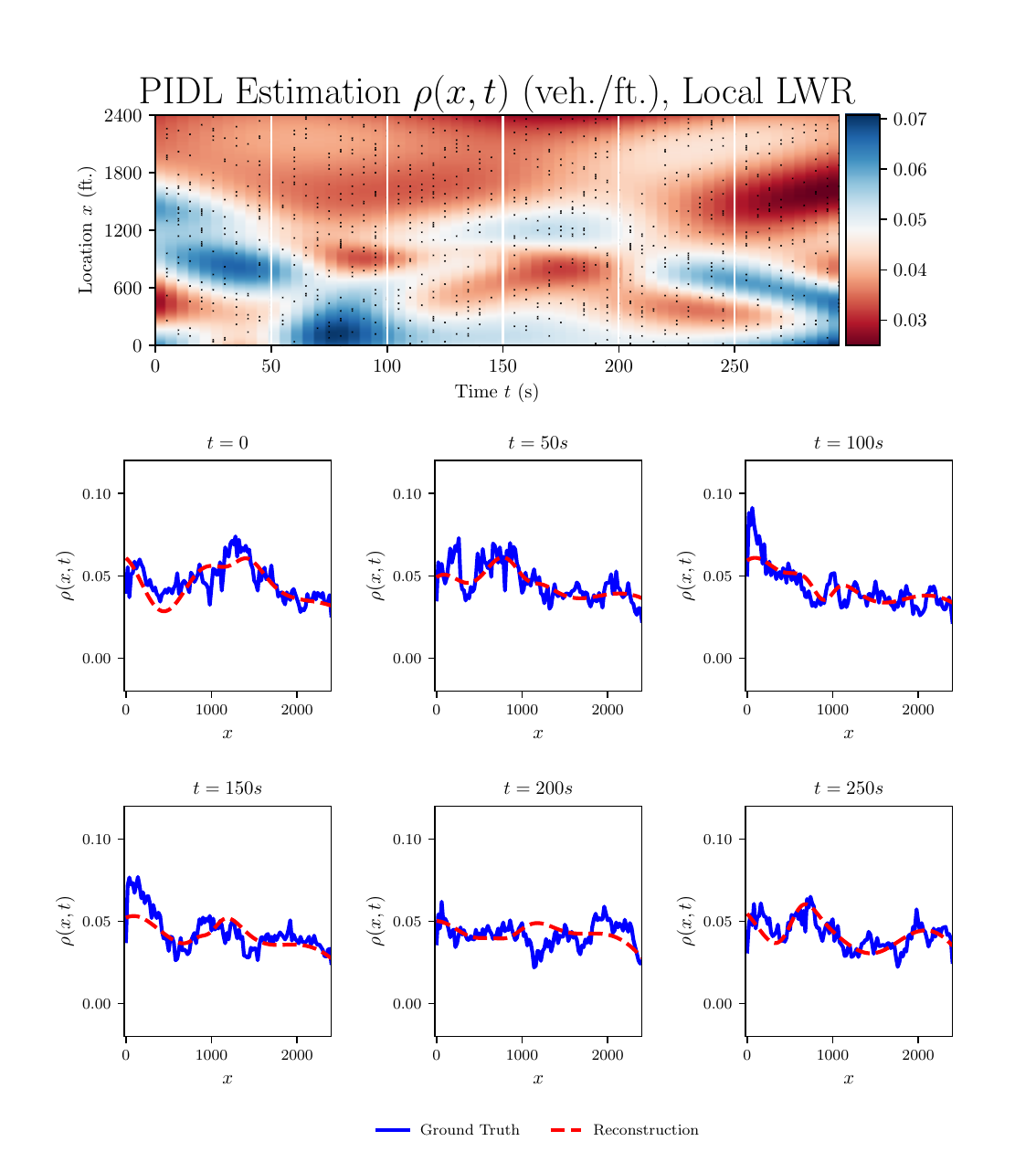}
    \caption{Local LWR}
    \label{fig:citysim_local_lwr}
     \end{subfigure}
     \hfill
     \begin{subfigure}[b]{0.33\textwidth}
         \centering
         \includegraphics[width=\textwidth]{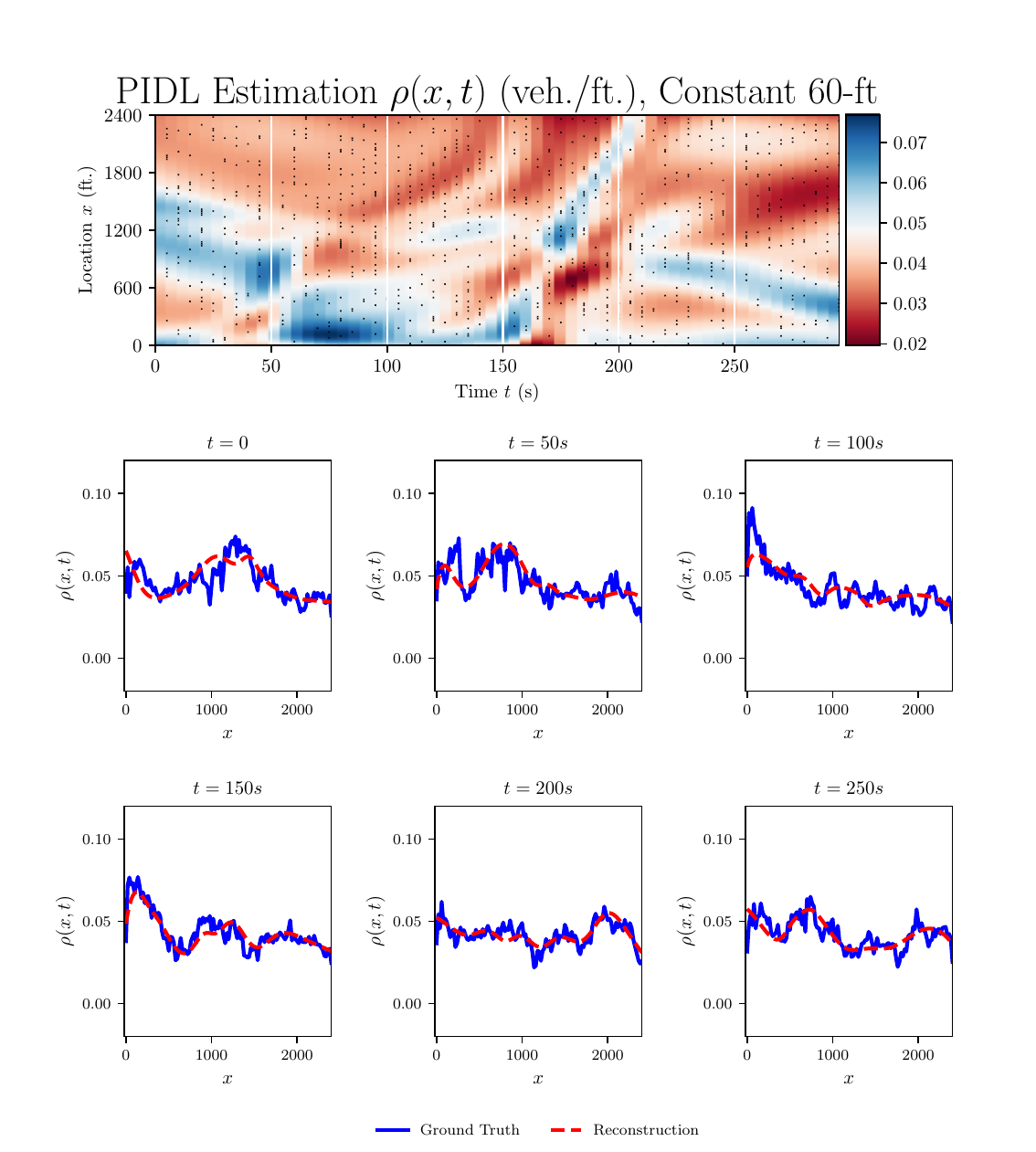}
    \caption{Constant Kernel, 60-ft}
    \label{fig:citysim_constant_kernel_60}
     \end{subfigure}
    \hfill
     \begin{subfigure}[b]{0.33\textwidth}
         \centering
         \includegraphics[width=\textwidth]{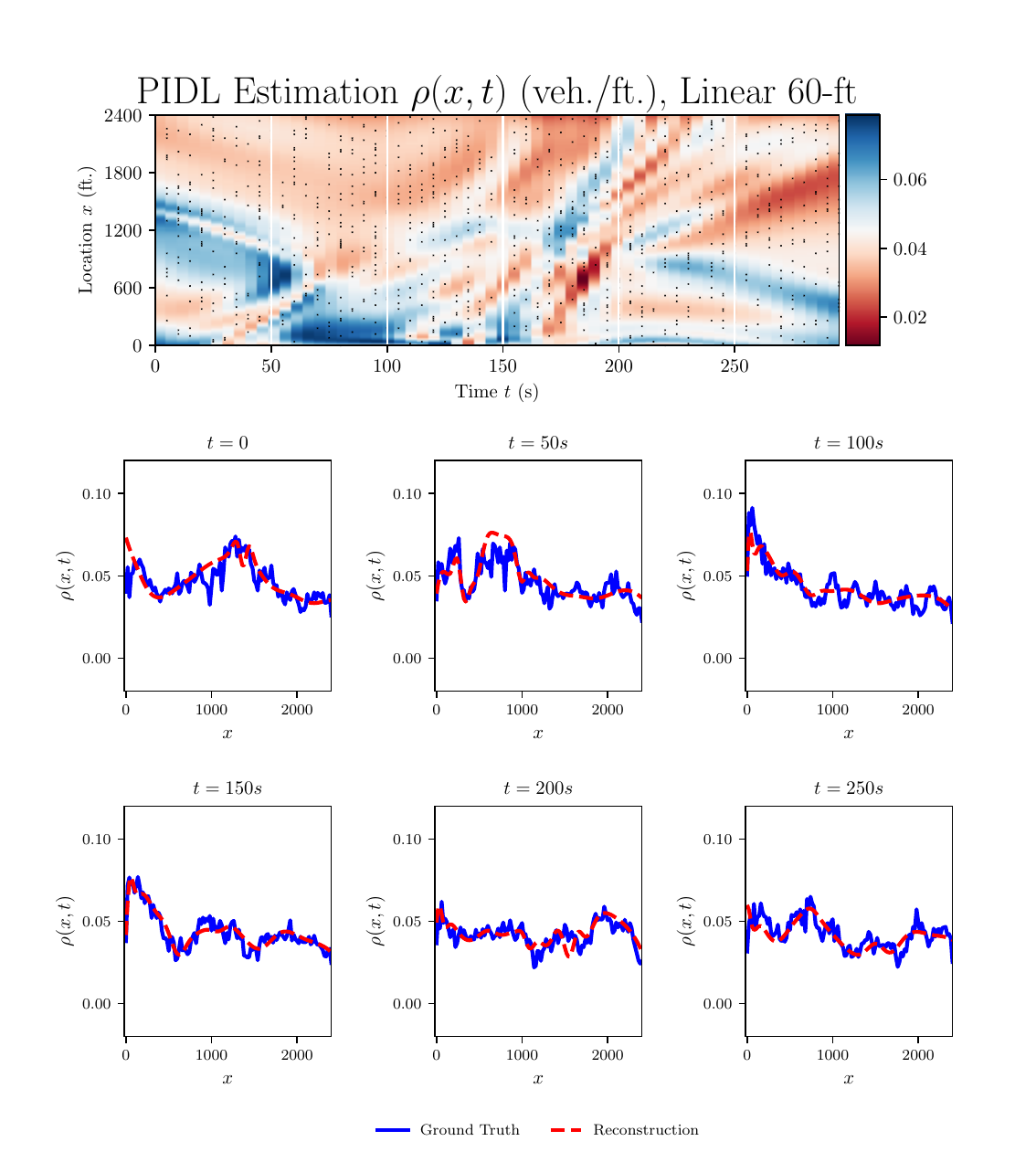}
    \caption{Linearly Decreasing Kernel, 60-ft}
    \label{fig:citysim_linear_kernel_60}
     \end{subfigure}
    \caption{CitySim Reconstruction with Local and Nonlocal LWR incorporated PIDL}
    \label{fig:citysim_reconstruction}
\end{figure*}

Fig.~\ref{fig:citysim_reconstruction} presents the traffic density reconstruction results achieved by using local LWR and $60$-ft fixed-length kernels (corresponds to the models 1 to 3 in Table~\ref{tab:citysim_I80_density_nonlocal}). 

Compared to the reconstruction result with local LWR in Fig.~\ref{fig:citysim_local_lwr}, the incorporation of nonlocal physics improved the speed reconstruction results. Observe that at $t = 150, 200, 250s$ snapshots in Figs.~\ref{fig:citysim_constant_kernel_60} and \ref{fig:citysim_linear_kernel_60}, where the proposed approach improved the reconstruction outcome of traffic density as compared to Fig.~\ref{fig:citysim_local_lwr}. In the scope of this extensive research, it has been observed that all the PIDL neural networks that incorporate nonlocal physics surpass the performance of the benchmark model, which is based on the local LWR PDE. The effectiveness of nonlocal physics is clearly manifested in the improved accuracy and robustness of the traffic state estimation tasks, underscoring the value of this approach in advancing the field of traffic flow modeling and prediction.

\begingroup
\setlength{\tabcolsep}{6pt} 
\renewcommand{\arraystretch}{1.4} 
    \begin{table*}[htbp]
        \caption{Performance Analysis of Nonlocal PIDL using CitySim Data}
        \begin{center} 
            \begin{tabular}{|p{0.5cm}|p{2.2cm}|p{3.0cm}|p{1.3cm}|p{3.5cm}|p{2.0cm}|}\hline
                & \textbf{Model Type} & \textbf{Look-ahead Kernel} & \textbf{Variable Length} & \textbf{Convolution Window} & \textbf{Relative $\mathcal{L}_2$ Error (\%)}\\\hline
                \textbf{1} & Local LWR    & $N/A$               & $N/A$ & $N/A$ & 17.01\\\hline
                \textbf{2} & Nonlocal LWR & Constant            & No & 60-ft & 15.34\\\hline
                \textbf{3} & Nonlocal LWR & Linearly decreasing & No & 60-ft & 15.17 \\\hline
                \textbf{4} & Nonlocal LWR & Constant            & No  & 100-ft & 15.40 \\\hline
                \textbf{5} & Nonlocal LWR & Linearly decreasing & No  & 100-ft & 15.57 \\\hline
                \textbf{6} & Nonlocal LWR & Constant            & Yes & Maximum length: 100-ft & 15.80 \\\hline
                \textbf{7} & Nonlocal LWR & Linearly decreasing & Yes & Maximum length: 100-ft & 15.73 \\\hline             
            \end{tabular}
            \label{tab:citysim_I80_density_nonlocal}
        \end{center}
    \end{table*}
\endgroup

\section{CONCLUSION} \label{sec:conc}

This study extended the Physics-Informed Deep Learning (PIDL) paradigm by integrating it with a nonlocal formulation of the Lighthill-Whitham-Richards (LWR) flow conservation law for traffic state estimation (TSE). Alongside employing fixed-length convolution kernels, we have also introduced variable-length kernels, addressing the `thick' upper boundary issue typically encountered in nonlocal models. The validation of our approach was carried out using real-world data from the Next Generation Simulation (NGSIM) and CitySim datasets. The results were promising, exhibiting improved accuracy in traffic state estimation compared to standard methods. Future research will explore the implementation of the variable-length kernel, where the window size is directly tied to the local density value, further refining the accuracy and reliability of our model.


\bibliographystyle{unsrt}
\bibliography{nonlocal} 
\end{document}